\documentclass[review]{elsarticle}

\usepackage{hyperref}

\usepackage{multirow}
\usepackage{dsfont}
\usepackage{amsmath}
\usepackage{algorithmic}
\usepackage{algorithm}
\usepackage{subfigure}

\usepackage{float}  
\usepackage{lipsum}
\makeatletter
\newenvironment{breakablealgorithm}
{
	\begin{center}
		\refstepcounter{algorithm}
		\hrule height.8pt depth0pt \kern2pt
		\renewcommand{\caption}[2][\relax]{
			{\raggedright\textbf{\ALG@name~\thealgorithm} ##2\par}%
			\ifx\relax##1\relax 
			\addcontentsline{loa}{algorithm}{\protect\numberline{\thealgorithm}##2}%
			\else 
			\addcontentsline{loa}{algorithm}{\protect\numberline{\thealgorithm}##1}%
			\fi
			\kern2pt\hrule\kern2pt
		}
	}{
		\kern2pt\hrule\relax
	\end{center}
}
\makeatother

\renewcommand{\algorithmicrequire}{\textbf{Input:}}

\journal{Journal of \LaTeX\ Templates}

\begin{document}

\begin{frontmatter}

\title{SGED: A Benchmark dataset for Performance Evaluation of Spiking Gesture Emotion Recognition}

\author[addressA,addressB,addressC,addressD]{Binqiang Wang}
\ead{wangbinqiang@ieisystem.com}

\author[addressB,addressC,addressD]{Gang Dong\corref{corrAuthor}}
\ead{donggang@ieisystem.com}

\author[addressB,addressC,addressD]{Yaqian Zhao\corref{corrAuthor}}
\cortext[corrAuthor]{Corresponding author}
\ead{zhaoyaqian@ieisystem.com}

\author[addressB,addressC,addressD]{Rengang Li}
\ead{lirengang.hsslab@gmail.com}

\author[addressE]{Lu Cao}
\ead{lu.cao@intel.com}

\author[addressB,addressC,addressD]{Lihua Lu}
\ead{lulihua@ieisystem.com}

\address[addressA]{Shandong Massive Information Technology Research Institute, Jinan, China,\\ }
\address[addressB]{State Key Laboratory of High-end Server \& Storage Technology, Beijing, China,\\}
\address[addressC]{Inspur (Beijing) Electronic Information Industry Co., Ltd., Beijing, China,\\}
\address[addressD]{Inspur Electronic Information Industry Co., Ltd., Jinan, China\\}
\address[addressE]{Intel Labs China, Beijing, China}

%

\begin{abstract}
In the field of affective computing, researchers in the community have promoted the performance of models and algorithms by using the complementarity of multimodal information. However, the emergence of more and more modal information makes the development of datasets unable to keep up with the progress of existing modal sensing equipment. Collecting and studying multimodal data is a complex and significant work. To supplement the challenge of partial missing of community data, we collected and labeled a new homogeneous multimodal gesture emotion recognition dataset based on the analysis of the existing data sets. This data set complements the defects of homogeneous multimodal data and provides a new research direction for emotion recognition. Moreover, we propose a pseudo dual-flow network based on this dataset, and verify the application potential of this dataset in the affective computing community. The experimental results demonstrate that it is feasible to use the traditional visual information and spiking visual information based on homogeneous multimodal data for visual emotion recognition. The dataset is available at \url{https://github.com/201528014227051/SGED}
\end{abstract}


\end{frontmatter}

\section{Introduction}
\label{sec:1}
Multimodal emotion recognition aims to enhance the stability and accuracy of emotion recognition by using the complementarity of multimodal information \cite{wang2022non}. The existing multimodal emotional tasks mostly include image, text, voice and other information \cite{zadeh2016mosi}. Through the interactive fusion of these information, we can make up for the lack of single mode data in some aspects of information. Just as the human five senses can better perceive the external world through cooperation, the collaborative fusion of multimodal data also improves the performance of emotion recognition tasks. It also provides technical reference for other fields based on multimodal information processing, such as matching, tracking, and brain imaging classification \cite{ning2021disentangled,jiang2019context,li2023unsupervised}.
\par
\begin{figure}
	\centering
	\scalebox{0.8}{\includegraphics[width=\textwidth]{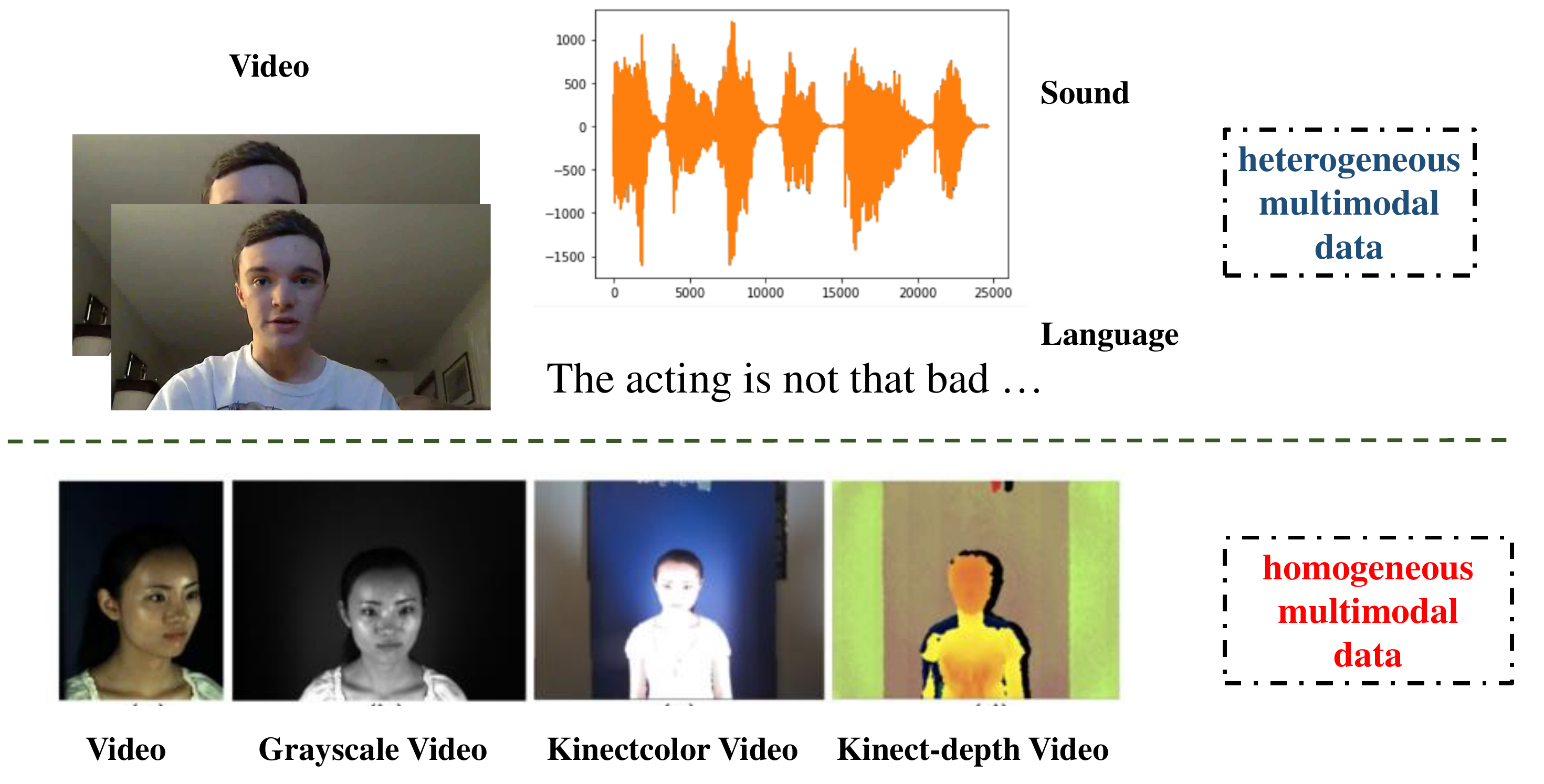}}
	\caption{The examples of different multi-modal data.}
	\label{intro_example}
\end{figure}

 
According to different sources of multimodal information, this paper classifies multimodal data into two categories: homogeneous multimodal data and heterogeneous multimodal data. Homogeneous multimodal data refers to data from unified information sources, such as visual effects from different angles or devices \cite{li20224dme}. Heterogeneous multimodal data refers to data from different information sources, such as multimodal data from vision and sound \cite{lu2019sound}. Due to the characteristics of heterogeneous multimodal data from different information sources, the data are more complementary, but the data form is different, so the processing mode between them is different. Homogeneous multimodal data generally come from different forms of the same information source, such as the combination of the popular ordinary RGB video information and depth camera, or the data acquisition form different from the traditional single RGB video \cite{li20224dme}. The research goal of homogeneous multimodality is to mine the task-related information contained in the data as much as possible from multiple angles under the same information source. Some examples is shown in Fig. \ref{intro_example} to show the difference between heterogeneous multimodal data and homogeneous multimodal data. It is worth noting that the information after homogeneous multimodal fusion can also be used as a single input in heterogeneous multimodal.
\par 
The information processing source of this paper is data from vision. Traditional video data includes videos containing a sequences of ordered pictures. Video can capture dynamic information relative to pictures, and can better predict tasks by capturing the context of time. Video from a single data source can only record information from one angle, and there is a visual dead angle. In order to record scene data more comprehensively, some researchers proposed to use multiple camera data and then model 3D data for emotion recognition \cite{li20224dme}. From another perspective, this paper fully exploits the potential of vision, records data of different format, and provides algorithms. Visual transmission is more about posture information. Some scholars have made relevant preliminary research on dynamic posture, but this paper has carried out more systematic exploration and analysis in the field of emotion recognition.
The performance of ordinary cameras in capturing scenes will decline under low illumination, fast motion, high dynamic range, etc. This decline in the performance of the input will directly affect the accuracy of subsequent tasks. The event flow data can just overcome these defects and provide a more robust recognition result.
\par 
In order to fully capture the dynamic information contained in the gesture, in addition to the ordinary video frame data, another modal data form used here is event stream data. The feature of event stream data is that it can capture the dynamic information in the picture, and only record the position and polarity of changes in the picture \cite{lichtsteiner2008128}. Moreover, the time resolution of event stream data recording is generally very high, and the dynamic range is larger. Considering these characteristics, we decided to adopt the homogeneous multimodal form of video frames and event streams for emotion recognition research. If two different devices are used, the problem of scene registration should also be considered, because the location of different devices is different. In order to avoid this difficulty, we adopted a collection device that can output video frames and event stream data at the same time, avoiding the error caused by the registration process from the hardware level \cite{brandli2014240}.
\par
The emotion recognition technology based on pure vision can be applied to situations where other modal information is not available, such as when the distance is relatively long and the surrounding noise is particularly large, only visual information is effective. Sign language is one of the most important means of information exchange without voice mode, especially for the hearing-impaired. However, the learning of professional sign language needs huge manpower, so it is positioned as ordinary daily communication gestures at the application level. That is, gestures that can be distinguished by people's common sense without special sign language learning can fill the gap in the field of multimodal emotion recognition data.
\par 
By analyzing the data characteristics of the two modes, the video frame data contains more semantic information, while the event stream data contains more dynamic information. Artificial neural network can effectively model semantic information, while spiking neural network also has natural advantages for event flow modeling. So the main contributions of this paper are as follows:
\begin{itemize}
	\item A new multimodal emotion recognition data paradigm is proposed, which includes video frames and event streams data collection and a labeling algorithm to find sample split index of stream data according to the annotation of video frames.
	\item Based on the characteristics of the data, the emotion recognition dataset based on dynamic gesture information is constructed. As far as we know, this is the first multimodal dataset containing emotion information posture in the community.
	\item we propose a new pseudo-double-stream network, which can be input in two streams: video frame data is processed by artificial neural network, and event stream data is processed by spiking neural network. The experimental results validate the effectiveness of multi inputs.
\end{itemize}

\par 

\section{Related Work}
\label{sec:2}
The organization form of event stream is to record whether the brightness of the scene changes and the location of the change by imitating the mechanism of retina periphery, which is different from the regular cumulative exposure of the general camera for the whole scene. The characteristics of event data are asynchronous and sparse, which can provide input for real-time applications with low latency \cite{zhu2022ultra}. Here, we first describe the relevant contents of the event dataset, and then introduce the algorithms and applications related to the event data.
\subsection{Event Dataset}
\par 
In order to obtain event data, the existing popular data collection devices include DVS (Dynamic Vision Sensor) \cite{lichtsteiner2008128}, DAVIS (Dynamic and Active-pixel VIsion Sensor) \cite{brandli2014240}, ATIS (Asynchronous Time-based Image Sensor) \cite{simon2016event}, and CeleX \cite{guo2017live}. From the perspective of equipment availability, DVS and DAVIS are relatively good choices. Of the two, DVS can only record pure event stream data, while DAVIS can also obtain corresponding video frame data with rich semantic information. Therefore, DAVIS (specifically, DAVIS346) is selected as the acquisition device in this paper.

Due to the inherent dynamic characteristics of gesture, dynamic vision camera has been applied to vision tasks for a long time. Mueggler \emph{et al.} proposed a simulator of event data that can be used for multiple tasks, which can provide data for the research of multiple tasks, such as pose estimation, visual odometry, and SLAM (Simultaneous Localization And Mapping) \cite{mueggler2017event}. However, this data is different from the real world data due to the simulation characteristics. So more researchers use physical cameras to collect event data. Most of the existing event data sets are captured using DVS \cite{lungu2017live,chen2019flgr,vasudevan2020introduction}, while the data sets captured using DAVIS are still few. Scheerlinck \emph{et al.} proposed to use DAVIS to record color event data sets, but the shooting method is to shoot static scenes by shaking the camera, and the focus is not gestures \cite{scheerlinck2019ced}. 
\par 
Lungu \emph{et al.} proposed a dataset containing three gestures: rock, scissors and paper \cite{lungu2017live}. In the same year, Amir \emph{et al.} released DvsGesture, the categories include hand waving (both arms), large straight arm rotations (both arms, clockwise and counterclockwise), forearm rolling (forward and backward), air guitar, air drums, and an ``Other" gesture \cite{amir2017low}. Subsequently, Chen \emph{et al.} released a dataset called Neuro ConGD Dataset, the gestures include beckoning, fingersnap, ok, push-hand (down, left, right, up), rotate-outward, swipe (left, right, up), tap-index, thumbs-up, zoom (in, out) \cite{chen2019flgr}.

Maro \emph{et al.} published a dataset, which is designed for the gestures controlling a smartphone, so when collecting data, one hand holding the smartphone while the other hand performs the movement \cite{maro2020event}. In the same year, the SL-Animals-DVS dataset was released, which focuses on using gestures to convey different animal information \cite{vasudevan2020introduction}. The datasets mentioned above are all single-mode DVS or ATIS of event data, and the information and emotion transmitted are not specifically targeted. The new data set proposed in this paper makes up for these defects in the field. The relevant information of the above datasets is shown in Table \ref{datasets_list}. It can be seen that this paper is not only relatively larger in data scale, but also more homogeneous video frame data in data mode than other datasets. The proposed SGED (Spiking Gestures Emotion Dataset) is also presented in Table \ref{datasets_list}, and the details progress of constructing the dataset will be introduced in Section \ref{dataset_section}

\begin{table}[]
\centering
\caption{Descriptions of different datasets related to gestures}
\label{datasets_list}
	\begin{tabular}{|c|c|c|c|c|}
		\hline
		\begin{tabular}[c]{@{}c@{}}Dataset Name/\\ Contributor\end{tabular} & Device   & \begin{tabular}[c]{@{}c@{}}Category \\ number\end{tabular} & \begin{tabular}[c]{@{}c@{}}Sample \\ number\end{tabular} & \begin{tabular}[c]{@{}c@{}}Spatial \\ resolution\end{tabular} \\ \hline
		ROSHAMBO17 \cite{lungu2017live}                                                         & DVS      & 3                                                          & 5 million                                                & 64$\times$64                                                         \\ \hline
		DvsGesture \cite{amir2017low}                                                         & DVS128   & 10+1(Other)                                                & 1342                                                     & 128$\times$128                                                       \\ \hline
		Neuro ConGD Dataset \cite{chen2019flgr}                                                & DVS      & 16+1(blank)                                                & 2040                                                     & 128$\times$128                                                       \\ \hline
		Maro and Benosman \cite{maro2020event}                                             & ATIS     & 6                                                          & 1621                                                     & None                                                          \\ \hline
		SL-Animals-DVS \cite{vasudevan2020introduction}                                                      & DVS      & 19                                                         & 1100                                                     & None                                                          \\ \hline
		SGED                                                                & DAVIS346 & 9+1(Other)                                                 & 2500                                                     & 346$\times$260                                                       \\ \hline
	\end{tabular}
\end{table}
\subsection{Algorithms and Applications}
\par 
The processing of event data can be divided into two steps. The first step is to encode the event data, and then apply it on the basis of coding. Many scholars have begun to explore various applications based on event data, including but not limited to classification, recognition, detection, and tracking \cite{cohen2015event,mitrokhin2018event}. Innocenti \emph{et al.} proposed a temporal binary representation method to encode event stream data. The time stream sequence with fixed spatial position is treated as a binary number, and then converted to decimal for subsequent operation \cite{innocenti2021temporal}. Messikommer \emph{et al.} proposed a general asynchronous sparse convolution structure based on events, which can fully utilize the sparsity and asynchrony of event data \cite{messikommer2020event}.

Lagorce \emph{et al.} proposed a hierarchy of event-based time-surfaces to extract informative features from event data for pattern recognition \cite{lagorce2016hots}. In order to match with the hardware, Tapiador \emph{et al.} conduct simulation of the hierarchy of time-surfaces for event based gesture recognition on FPGA (Field Programmable Gate Arrays) \cite{tapiador2020event}. In order to help the elderly or people with visual impairment use mobile phones, Maro \cite{maro2020event} proposed a low-energy solution with dynamic background suppression based on event cameras, which uses the computing power provided by mobile phones to control mobile phones through recognizing basic gestures. Ceolini \emph{et al.} uses multimodal input including event data and EMG (Electromyography) signal to combine arm and gesture information to determine category of hand-gesture \cite{ceolini2020hand}. In order to complete the real-time event stream data processing, Linares \emph{et al.} gives a FPGA implementation scheme based on event data filtering and feature extraction \cite{linares2019low}. Muthusamy \emph{et al.} has designed a robot equipped with DVS, so that the robot finger can overcome the uncertainty of different light changes and jitters in the process of grasping and stabilizing objects \cite{muthusamy2020neuromorphic}.

SNN (Spiking Neural Network) has better biological interpretability and can directly deal with event data, so it is widely used in relevant research. Liu \emph{et al.} design a spiking neural network using motion information for event based action recognition,  including motion perception, motion pooling, spatial pooling, and classifier \cite{liu2021event}. In order to achieve the classification and detection of low energy consumption, Kugele \emph{et al.} proposed a hybrid ANN (Artificial Neural Network) and SNN scheme. The front part of the network uses SNN to process the input event data, and the back end uses ANN to obtain the final task output \cite{kugele2022hybrid}. With the popularity of Transformer, Zhang \emph{et al.} added event data to it to form a spiking Transformer for target tracking \cite{zhang2022spiking}.

In order to build three-dimensional information, Andreopoulos \emph{et al.} uses two DAVIS cameras, one left and one right, to build a low power, high throughput, fully event-based stereo system \cite{andreopoulos2018low}. In addition, in order to reduce the energy consumption including matrix vector multiplication calculation, resistive memory array is also a possible choice \cite{wang2023echo}. It is also an important application to add visual event data to autonomous driving system \cite{maqueda2018event,chen2020event}.
Although the research scope of this paper is visual event data, by the way, some scholars have studied event-based silicon cochlea \cite{li2012real}.
\begin{table}[]
\centering
\caption{Descriptions of different emotional gestures}
\label{dataset_description}
\scalebox{0.85}{
	\begin{tabular}{|c|c|c|}
		\hline
		\begin{tabular}[c]{@{}c@{}}affective \\ disposition\end{tabular} & \begin{tabular}[c]{@{}c@{}}emotional \\ gestures\end{tabular} & Gesture details                                                                                                                                                                                                    \\ \hline
		\multirow{2}{*}{Neutral}                                         & ok                                                            & \begin{tabular}[c]{@{}c@{}}The thumb and index finger form a circle, \\ and the remaining fingers naturally open\end{tabular}                                                                                      \\ \cline{2-3} 
		& hello                                                         & \begin{tabular}[c]{@{}c@{}}Combine the five fingers, extend the palm, \\ and raise the hand to the eyebrow\end{tabular}                                                                                            \\ \hline
		\multirow{2}{*}{Negative}                                        & no                                                            & Shake your head left and right                                                                                                                                                                                     \\ \cline{2-3} 
		& kill                                                          & The palm slides left and right in front of the neck                                                                                                                                                                \\ \hline
		\multirow{5}{*}{Positive}                                        & victory                                                       & \begin{tabular}[c]{@{}c@{}}Palm outward, \\ extend the index finger and middle finger in a 'V' shape\end{tabular}                                                                                                  \\ \cline{2-3} 
		& good                                                          & Thumbs up, other fingers clenched                                                                                                                                                                                  \\ \cline{2-3} 
		& yes                                                           & Nod up and down                                                                                                                                                                                                    \\ \cline{2-3} 
		& love                                                          & \begin{tabular}[c]{@{}c@{}}Stretch out the thumb, index finger, and the little finger, \\ the palm outward, and the thumb and index finger are in an 'L' shape\end{tabular}                                        \\ \cline{2-3} 
		& fighting                                                      & \begin{tabular}[c]{@{}c@{}}One hand or two hands, clench the fist and raise the hand, \\ with the center of the fist inward, \\ the forearm vertical to the ground, move down slightly, and then stop\end{tabular} \\ \hline
	\end{tabular}}
\end{table}
\section{Proposed Dataset}
\label{dataset_section}
\subsection{Basic Information}
To construct the multi-modal dataset containing both ordinary vision information and dynamic vision information, DAVIS346 (Dynamic and Active Pixel Vision Sensor) is used to capture the data. The temporal resolution of the DAVIS346 is 1 microsecond, which can capture the useful information lost by the ordinary camera. Especially, the DAVIS346 is able to collect more information when under backlight because of the high dynamic range of DAVIS346. Ten volunteers are involved in our experiments. For each volunteer, deliberately designed emotional gesture is conducted in different light conditions and body positions. There are totally nine categories: ok, hello, no, kill, victory, good, yes, love, and fighting. These emotion gestures are conveyed by gestures of body and the details are described in Table \ref{dataset_description}.
\subsection{Labeling Process}
The labeling process of multimodal data based on video frames and event streams is different from that of traditional multimodal data. Take the traditional multimodal annotation of video and audio as an example. It is assumed that video and audio are synchronized. When the original video is labeled, the same timestamp can be directly used to label the audio. However, this is not the case for video frame data and event stream data, which is caused by the difference in time resolution between the two. The output format of DAVIS346 is not a traditional video, but a simple video frame, which is saved in the form of compression and segmentation; On the other hand, the event stream is also saved in a tensor way, and cannot be directly aligned with the video frame for visualization. Due to the difference in time resolution, the number of frames in video frame data and the number of events in event stream data differ greatly. In order to complete the labeling of the collected data, this paper proposes a binary search algorithm based on the scaling factor to align two different types of data, so as to obtain the labeling serial number of the event flow data. The specific labeling process is shown in Algorithm \ref{alg}. 

\begin{breakablealgorithm}
	\caption{Scaling Binary Search Algorithm}
	\label{alg}
	\begin{algorithmic}[1]
		\begin{footnotesize} 
		\REQUIRE The time stamp corresponding to the video tag point, stored in a list with length $M-1$ (split into $M$ segments), denoted as $TagList$. Time information in event tensor data, stored in a list with length $N$, denoted as $TimeList$. Scaling factor $\alpha$.
\ENSURE Time index of different categories, stored in a list, denoted as $OutIndex$.
\renewcommand{\algorithmicrequire}{\textbf{Function}}
\REQUIRE{ScalingBinarySearch}($List$,$l$,$r$,$t$,$\alpha$):
\IF{$R$>=$L$}
\STATE $mid$ = int($l+$($r-l$)/2)
\IF{absolute value of $List$[mid]-$t$ $ is smaller than \alpha$}
\STATE return $mid$
\ELSIF{$List$[mid]$>t$}

\STATE return \textbf{ScalingBinarySearch}($List$,$l$,$List$[mid],$t$,$\alpha$)
\ELSE 
\STATE return \textbf{ScalingBinarySearch}($List$,$List$[mid],$r$,$t$,$\alpha$)
\ENDIF
\ELSE
\STATE return $-1$
\ENDIF
\renewcommand{\algorithmicrequire}{\textbf{End Function}}
\REQUIRE
\renewcommand{\algorithmicrequire}{\textbf{Function}}
\REQUIRE{FindPosition}($Tag$, $TimeList$):
\STATE $length$ $\gets$ obtain the length of $TimeList$
\IF{$Tag<$$TimeList$[1]} 
\STATE return $1$
\ENDIF
\IF{$Tag>$$TimeList$[$length$]} 
\STATE return $length$
\ENDIF
\STATE $\alpha$ = 1
\STATE $SearchTag$ = -1
\WHILE{$SearchTag$ = -1}
\STATE $SearchTag$ = \textbf{ScalingBinarySearch}($TimeList$, 1, $length$, $Tag$, $\alpha$)
\IF{$SearchTag$ = -1}
\STATE $\alpha$ = $\alpha$ + 1
\ENDIF
\ENDWHILE
\renewcommand{\algorithmicrequire}{\textbf{End Function}}
\REQUIRE
\FOR{$i=1$ to $M-1$}
\renewcommand{\algorithmicrequire}{\textbf{Function}}

\STATE $idx$ = \textbf{FindPosition}($TagList[i]$, $TimeList$)
\STATE $OutIndex$ $\gets$ $idx$
\ENDFOR
		\end{footnotesize}
	\end{algorithmic}
\end{breakablealgorithm}

\subsection{Statistical Analysis}
\begin{figure}
	\centering
	\scalebox{0.88}{\includegraphics[width=\textwidth]{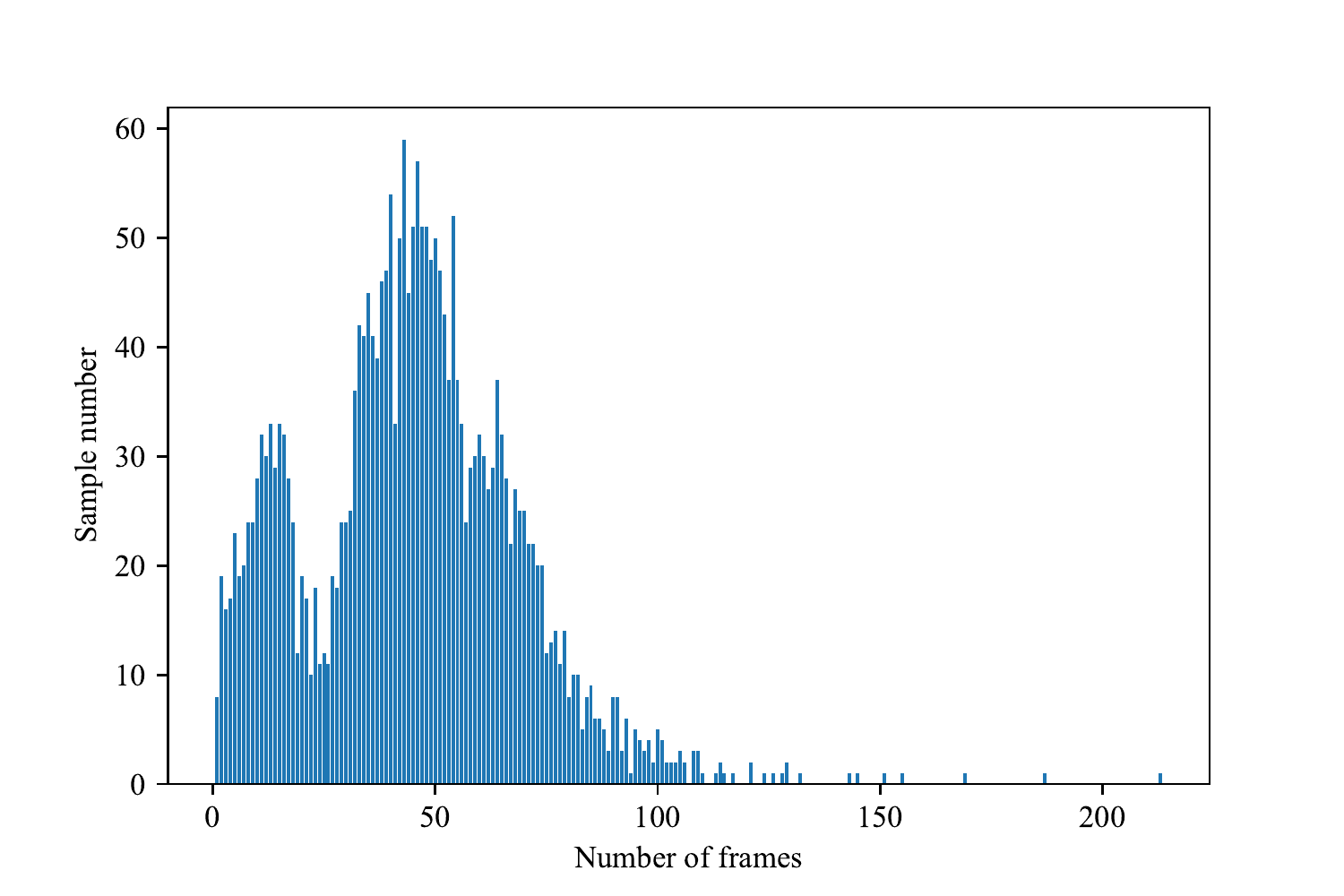}}
	\caption{Distribution of sample number and frame number of video frame data.}
	\label{frame_lens}
\end{figure}

To gain an intuitive understanding of the dataset as a whole, we conducted a simple statistical analysis of the dataset to find the prior information contained in it. The homogeneous multimodal dataset contains two parts of data, including video frame data and event stream data. For video frame data, the main parameter is the number of video frames in each sample. Here, we counted the number of samples containing different video frames, which are presented by histogram. As shown in Figure \ref{frame_lens}, it can be found that most of the sample video frames are concentrated in 0-100 frames. In this way, when designing the input of video frame module, we can use this 100 as a priori information to truncate the input video frame sequence.
\par
\begin{figure}
	\centering
	\scalebox{0.88}{\includegraphics[width=\textwidth]{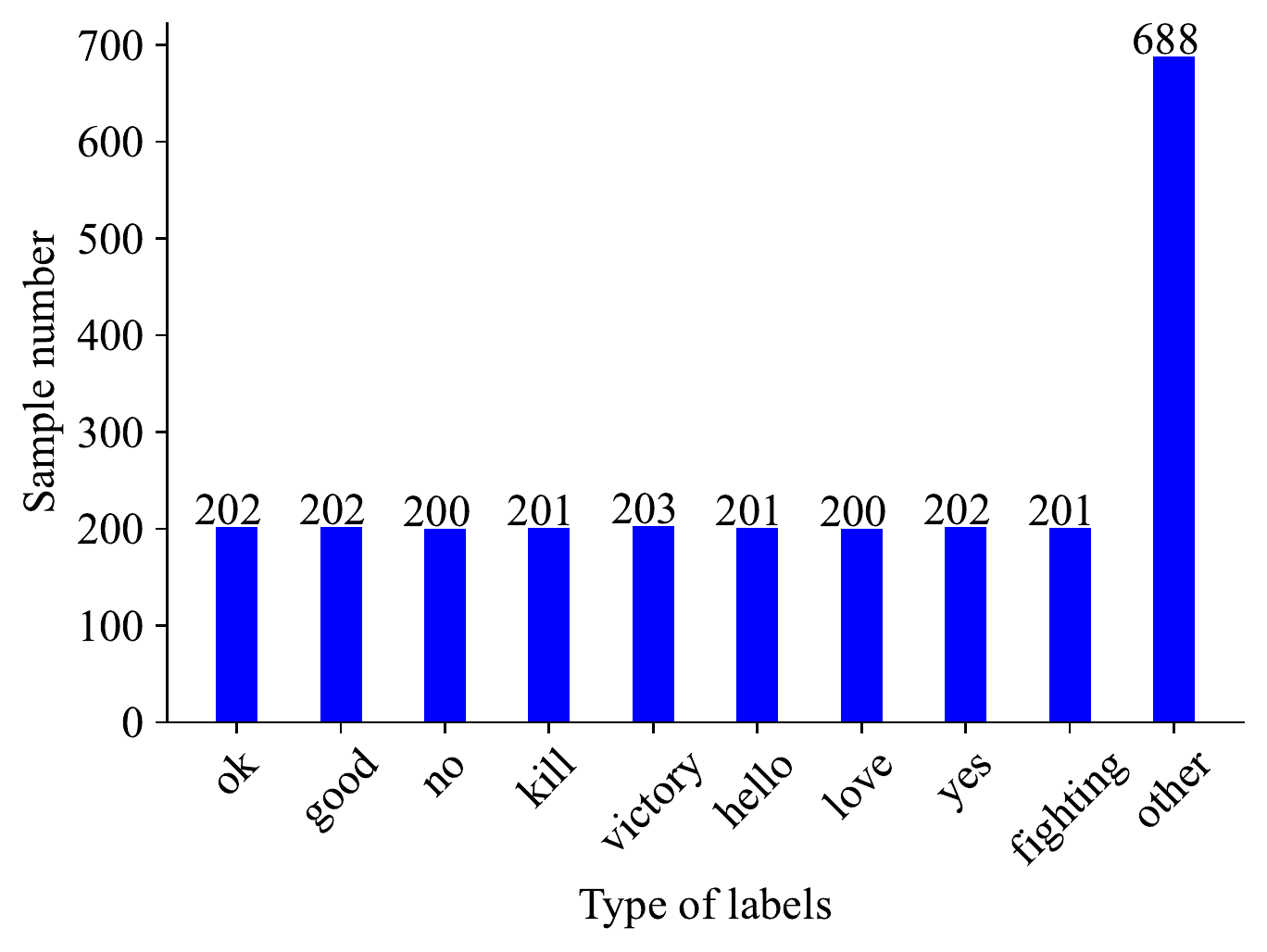}}
	\caption{Distribution of sample number and labels.}
	\label{imbalance_bar}
\end{figure}
\begin{figure}
	\centering
	\scalebox{0.88}{\includegraphics[width=\textwidth]{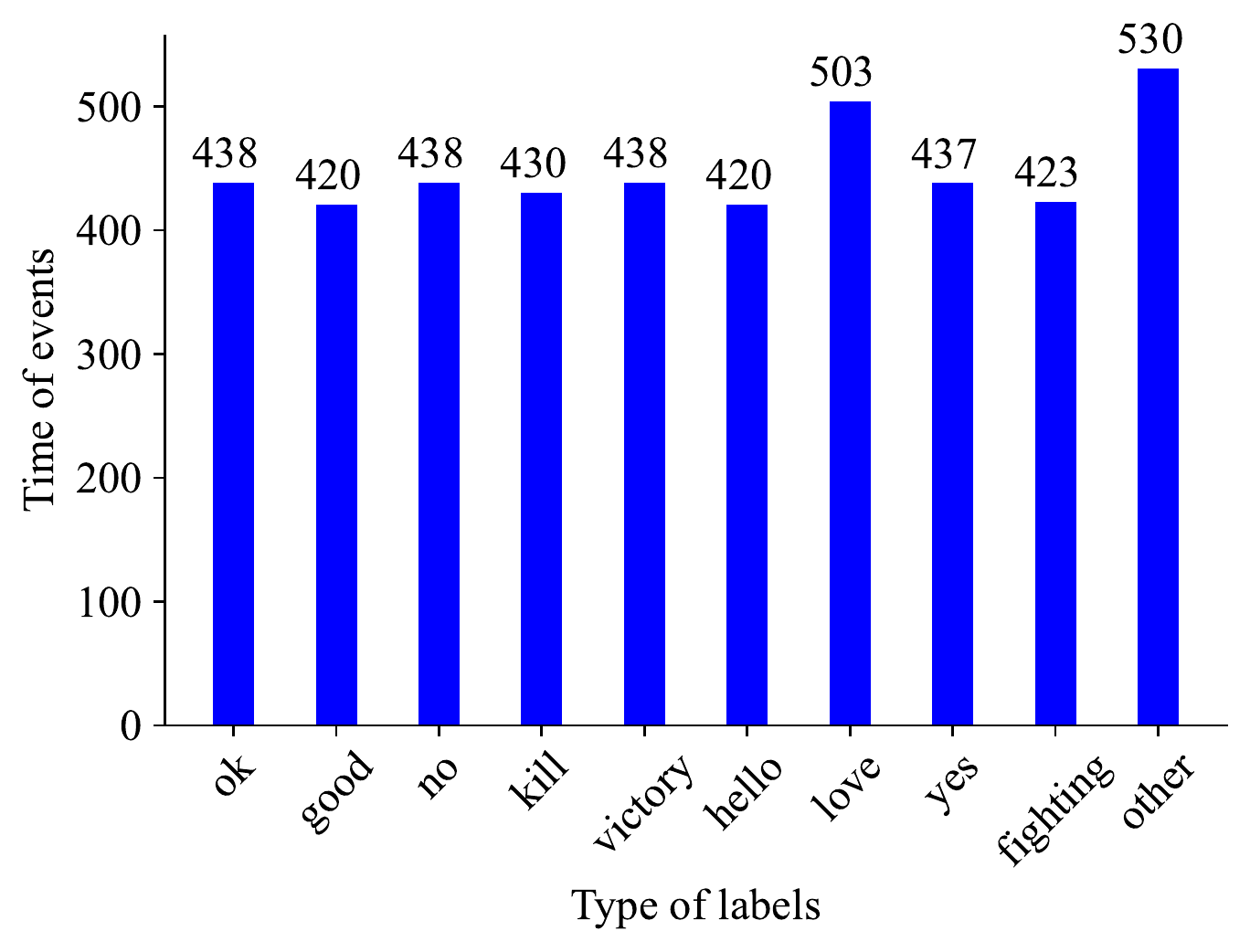}}
	\caption{Distribution of events' time summation of different categories (the unite of time is s(second)).}
	\label{event_time_bar}
\end{figure}

Although the data was collected according to the average category setting, in the labeling process, due to the setting of a redundant category, there is a phenomenon of category imbalance. As shown in Figure \ref{imbalance_bar}, the first nine categories are basically about 200 samples, while the tenth category is 688 samples. For the convenience of processing, the tenth sample was directly removed in the experimental part.
To illustrate the balance of sample data from a time length perspective, we have counted the duration of different sample event data. From Figure \ref{event_time_bar}, it can be seen that the overall duration of most categories of events is maintained at about 440s after adding on the entire data set. The gesture `love' takes a long time, probably because the gesture content is relatively complex, and the complex embodiment can be seen from the description in Table \ref{dataset_description}. In addition, the long time of `other' is mainly accumulated by quantity, so we will not pay special attention here.
\par
\begin{figure}
	\centering
	\scalebox{0.88}{\includegraphics[width=\textwidth]{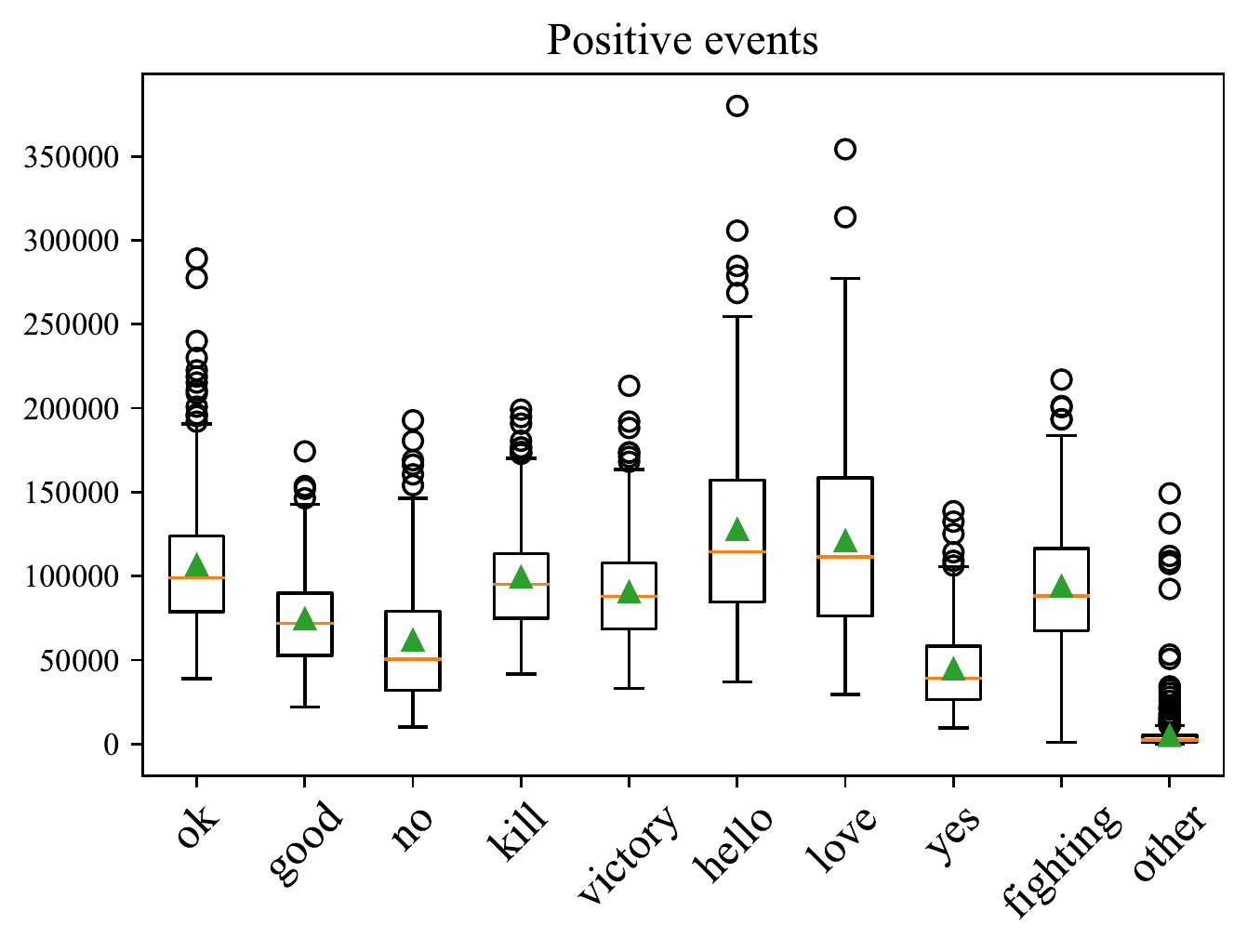}}
	\caption{Box diagram of the number of positive events.}
	\label{positive_box}
\end{figure}

\begin{figure}
	\centering
	\scalebox{0.88}{\includegraphics[width=\textwidth]{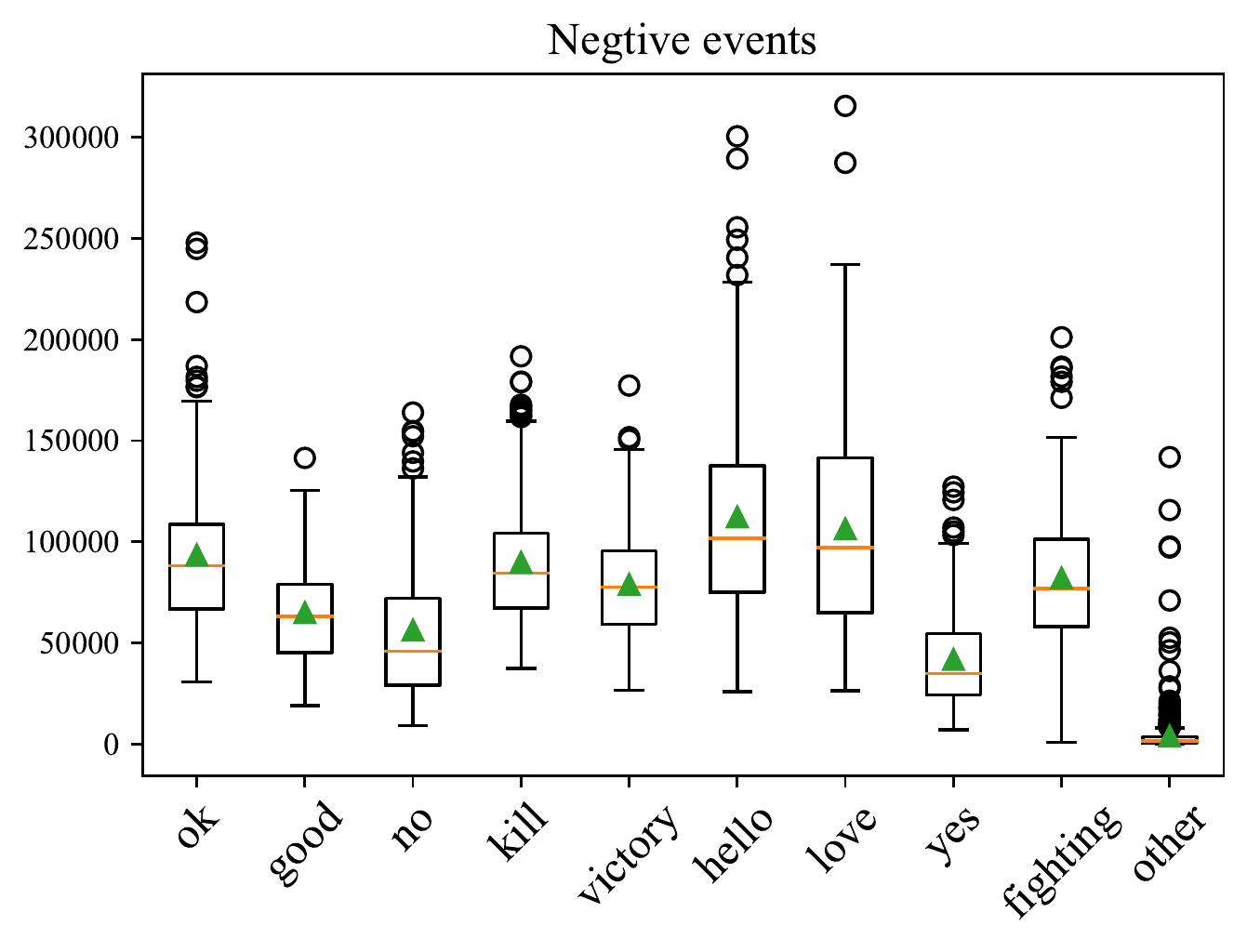}}
	\caption{Box diagram of the number of negative events.}
	\label{negative_box}
\end{figure}

 
To gain a deeper understanding of the distribution law of event data, we have made statistics on positive and negative events respectively. We made statistics on the number of positive and negative events contained in each sample in each category, and showed the mean and outlier information through the box diagram. As shown in Figures \ref{positive_box} and \ref{negative_box}, first of all, it can be found that the number of events (whether positive or negative) contained in different types of emotional postures is different. Secondly, we can predict the number of events contained in different emotional postures from the generation process of gestures. For example, the number of events contained in emotional postures containing complex movements (`love') is more than that contained in emotional postures containing only head movements (`yes' and `no'). Moreover, the distribution of positive and negative events is positively correlated, that is to say, the sample with a large number of positive events corresponds to a large number of negative events. This is because in the process of gesture operation, it basically belongs to rigid body motion, and the trigger of events is positive and negatives imultaneously in most conditions, with differences in location.

\begin{figure}
	\centering
	\scalebox{0.88}{\includegraphics[width=\textwidth]{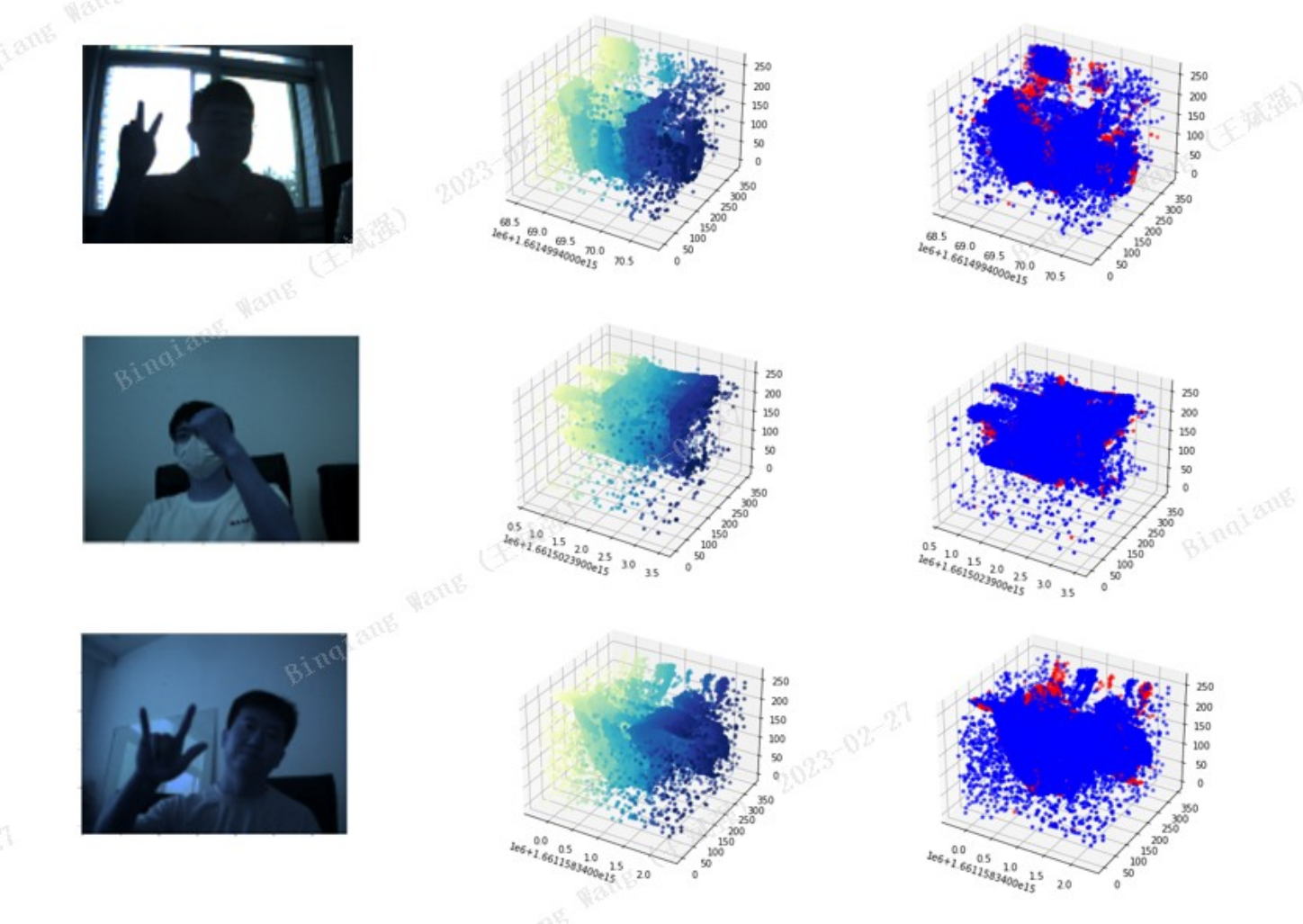}}
	\caption{Three examples from SGED.}
	\label{example_in_sged}
\end{figure}

To visually present the representation of data in the dataset, we have visualized the video frame data and corresponding event data. As shown in Figure \ref{example_in_sged}, the first column represents one of three video frame sequences with different poses. The second column represents the schematic diagram of the change of event stream data in two-dimensional space over time (from yellow to blue represents the beginning to end of the event). It can be seen that the place where the event is concentrated always appears in the place where the action occurs. The third column represents the distribution of positive and negative events along with events, where blue represents positive events and red represents negative events. It can be seen that positive events are concentrated in the space where the hand appears.
\par 
It is worth noting that considering that the video frame information involves privacy data such as human faces and postures, we conducted a questionnaire survey on volunteers before releasing the data. According to the questionnaire, we published video frame data for agreed volunteers. For other data, in order to facilitate subsequent scientific research, we provided corresponding video frame feature data.
\section{Methods}
\begin{figure}
	\centering
	\scalebox{0.35}{\includegraphics{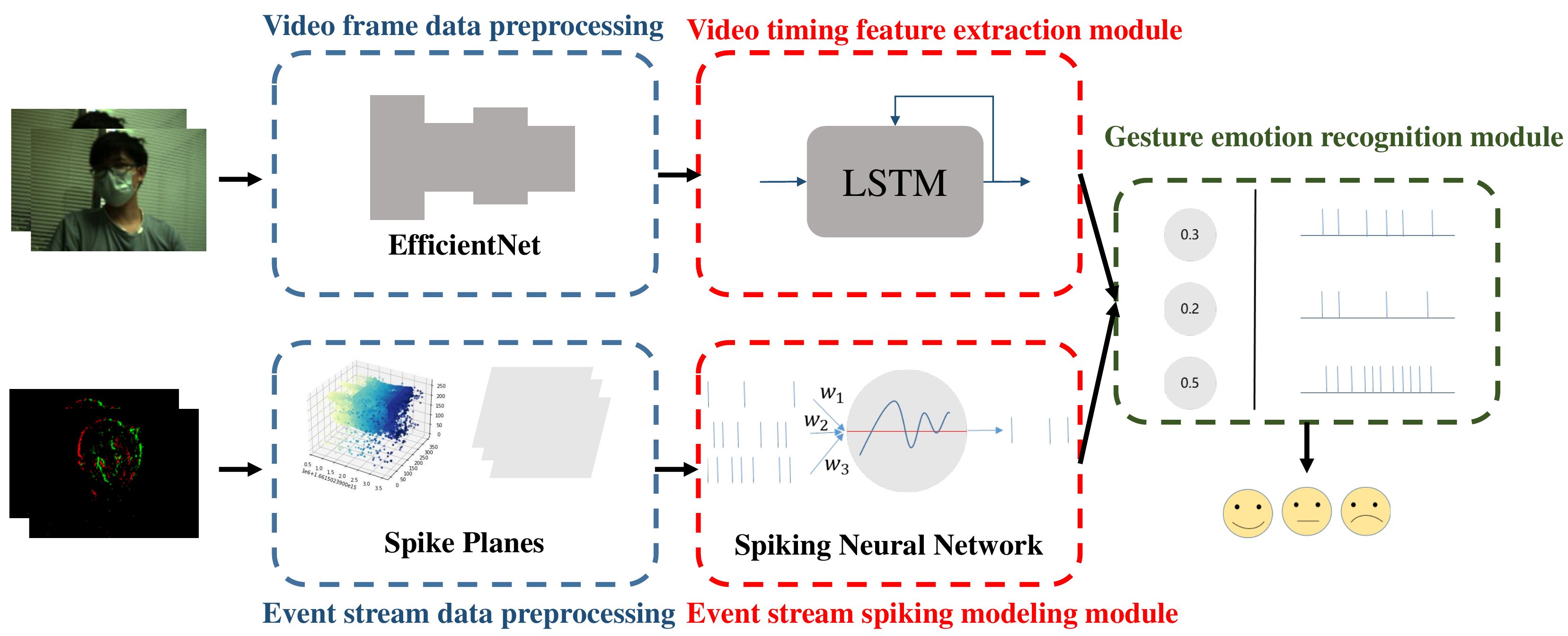}}
	\caption{The framework of the proposed method}
	\label{framework}
\end{figure}
\label{sec:3}
To achieve gesture emotion recognition based on homogeneous multimodal input data, we designed a model framework called homogeneous pseudo dual flow network. Here, the concept of dual flow networks \cite{wang2022high} is borrowed and extended to the homogeneous multimodal domain. This framework is mainly divided into the following parts: video frame data preprocessing, event stream data preprocessing, video timing feature extraction module, event stream spiking modeling module, and output module. The following will expand and introduce the contents of these modules.
\par 
For the symbolic representation of homogeneous multimodal data, we use $v_{i}, i=1,..,N$ to represent video frame data, where $N$ represents the number of video frames; $s_{j}, i=1,...,L$ represents a spike train, where $L$ represents the number of spikes. More specifically, each spike is represented by a vector, $(t,x,y,p)$, where $t$ represents the time at which the event occurred, $(x, y)$ represents the spatial location at which the event occurred, and $p$ represents the polarity at which the event occurred. The polarity can be divided into two cases, positive and negative, represented by 1 and 0. The goal of the model is to obtain output $\hat{y}$ based on the input data. Ideally, this $\hat{y}$ should be the same as the ground truth category $y$ that represents real gesture emotions.
\subsection{Video frame data preprocessing}
In this section, we introduce the preprocessing operations of video data. Because the obtained video sequence data is traditional RGB three-channel data, according to existing mainstream feature extraction methods, it is possible to directly extract three-dimensional features or first extract video frame features, and then use sequence modeling tools, such as LSTM, to model temporal features. In order to enhance the flexibility of subsequent model design and maximize respect for the privacy of volunteers without affecting subsequent model research, we adopt the second scheme here. For the extraction of video frame features, we choose the model, EfficientNet, from the perspective of efficiency. In this way, the video frame sequence eventually becomes a sequence composed of feature vectors, and the above process can be expressed as:
\begin{equation}
vp_{i} = f_{EN}\left ( v_{i} \right ) ,i=1,...,N,
\end{equation}
where $f_{EN}$ represents the feature extraction of EfficientNet, $vp_{i}$ is the feature vector of the video frame $v_{i}$, which is used as the input of the video timing feature extraction module.

\subsection{Event stream data preprocessing}
This section describes how to preprocess event data. The recording format of event data directly output by DAVIS346 is a four-dimensional vector sequence, as previously mentioned. This recording format is very similar to the recording format of point clouds (except for two-dimensional plane space, the third-dimensional space is a temporal dimension and the fourth-dimensional space records polarity information), so it can be processed using a similar encoding method for point clouds, but this can cause problems such as loss of spatial characteristics of the data, size mismatch, and so on. Therefore, the existing event-based processing algorithms try to map the sequence of events back to the original two-dimensional space, which is commonly referred to as the spike plane. The event can be restored to the two-dimensional coordinate position of the corresponding spike plane based on the spatial location of the event. After direct mapping, the original point cloud data is restored to spike plane sequences. At this point, there is a challenge that the number of spike planes is too large, which can lead to excessive subsequent computation. To solve this problem, researchers have proposed a solution to compress the spike plane. There are two main schemes available, one is to compress based on a fixed time interval, and the other is to compress based on the number of events. Due to the uncertainty of the occurrence of events, there may be situations where no events occur for a period of time depending on the time interval, resulting in an invalid spike plane (i.e., all data is zero). Therefore, compression based on a fixed number of events is used here.
\par 
Specifically, first, the number of all events included in a sample is counted, without distinguishing the polarity of the event. Then, according to a preset $K$, representing the number of dense spike planes after data preprocessing. It should be noted that due to two different polarity events, the dense spike plane is essentially a three-dimensional tensor with a third dimension of 2. To represent that the data is obtained by compressing the original spike plane sequence, it is called a dense spike plane. The numerical value in the dense spike plane is obtained by separately counting the number of corresponding positive and negative spike events. So far, the processing of the original event data stream has been completed, and dense spike plane sequence data has been obtained. The formal description of the above process is as follows:
\begin{equation}
sp_{1},...,sp_{K}=f_{dense}\left ( s_{j} \right ),j=1,...,L,
\end{equation}
where $f_{dense}$ denotes the compress from spike planes to dense spike planes, $sp_{1},...,sp_{K}$ are the dense spike planes. Note the for $s_{j}$, the channel number is 1 which means the positive and negative events are mixed. 
\subsection{Video timing feature extraction module}
In this section, we model temporal dependencies in video frame sequences. The output results obtained in the video frame data processing stage are processed separately for each frame, and the extracted feature vectors contain the semantic information of a single frame. In order to capture timing information between consecutive frames, it is necessary to model specifically. Here, the classic time series modeling model LSTM is used to provide a baseline. LSTM can control the compression of effective information from front to back through a gating mechanism, using the output data of the last step of LSTM as the temporal feature representation of the video. The timing modeling process of LSTM can be expressed as follows:
\begin{equation}
h_{last} = LSTM\left ( vp_1,...,vp_N \right ) ,
\end{equation}
where $h_{last}$ is the output of last hidden states of LSTM.
\par
It is important to note that LSTM is used here to verify the effectiveness of the complementarity of homogeneous multimodal data, and can be completely replaced by other temporal modeling tools, such as Transformer. 
\subsection{Event stream spiking modeling module}
This section focuses on modeling event data, which has unique spike characteristics that make it well-suited for spiking neural networks. This approach is commonly used for gesture recognition with event cameras. In this work, we employ a classic framework for modeling event flow data, with the key modification being the adjustment of the number of intermediate-layer neurons to match the output types \cite{amir2017low}. The spiking neural network topology includes standard convolution, pooling, and fully connected layers, which are used to spatially compress input data and extract meaningful features. The main difference in our approach is the configuration of the number of spiking neurons in the fully connected layer prior to the output layer.

The specific spiking neuron model uses LIF (Leaky Integrate-and-Fire) \cite{20230613559916}. An effective event flow feature extraction process can be expressed as:
\begin{equation}
	s_{DG} = DvsGesture \left ( sp_1,...,sp_K \right ) ,
\end{equation}
where $DvsGesture$ is the network structure in \cite{amir2017low}, $s_{DG}$ is the output spike array whose size is equal to the number of category. Note that $s_{DG}$ is after averaged with $K$, and direct used to compute the MSE loss.
\subsection{Gesture emotion recognition module}
This section describes how to synthesize the output of two different networks to obtain the final result. For spiking neural networks, we average pool the outputs of final spiking neurons output by the corresponding module (i.e. Event stream spiking modeling module) to reduce the dimension to the emotional category domain space. For the LSTM structure, we take the hidden state of the final step output as input and map it to the emotional output space through two fully connected layer with dropout rate 0.5. Due to the structure of the two branches involved, the output fusion method is directly used here.
\begin{equation}
 \hat{y} = s_{DG} + f_{o}\left( h_{last}\right),
\end{equation}
where $ f_{o}$ represents two fully connected layer with dropout rate 0.5. 

In addition, in experiments, we found that spiking neural networks need to use MSE losses to achieve good performance, while artificial neural networks are more effective in using cross entropy losses for this type of classification problem. Therefore, here, we balance the contribution of two different outputs by setting a hyperparameter and ultimately output the recognized gesture's emotional tendencies.

\section{Experiments}
\label{sec:4}
\subsection{Experimental Setup}
We will conduct experimental verification on the dataset released in this work. Because the goal of this paper is to identify the emotions involved in gestures, we ultimately abstract the task into a three class problem. According to the description in Table \ref{dataset_description} and the introduction to the SGED in the previous section, it can be found that the number of neutral and negative samples is close, while the number of positive samples is more than the other two combined. Thus we add class number to weight different class when computing cross entropy loss. This point reminds us that when carrying out model evaluation, we cannot simply measure the quality of the model based on accuracy, but should consider the imbalance of the sample.

The number of dense spike planes $K$ is set to 12. Adam is used to optimize the parameters. The comparison method used is the classic gesture recognition method \cite{amir2017low}. Because gesture emotion recognition can be seen as a classification task, the accuracy and confusion matrix are used to evaluate the model. In order to take into account sample imbalance in the evaluation of model performance, we also used weighted precision, weighted recall, and weighted F1 score as the evaluation criterion.

The split of data sets can have different effects on experimental results, so when we first released the dataset, we gave a variety of different methods for splitting the dataset and chose one of them to conduct the experiment. It is worth noting that due to the limitations on the amount of data currently released, the distribution of real data may differ from that of the collected dataset. In the future, this problem can be alleviated by collecting datasets that contain more samples.

\subsection{Experimental Results}

\begin{table}[]
\centering
\caption{Evaluation on SGED.}
\label{res_table}
	\begin{tabular}{|c|c|c|c|c|}
		\hline
		& Accuracy & \begin{tabular}[c]{@{}c@{}}Weighted\\ Precision\end{tabular} & \begin{tabular}[c]{@{}c@{}}Weighted\\ Recall\end{tabular} & \begin{tabular}[c]{@{}c@{}}Weighted\\ F1\end{tabular} \\ \hline
		DvsGesture & 63.4     & 66.0                                                         & 63.4                                                      & 60.8                                                  \\ \hline
		ours       & 64.5     & 64.1                                                         & 64.5                                                      & 64.1                                                  \\ \hline
	\end{tabular}
\end{table}
The experimental results on the SGED are shown in the Table \ref{res_table}. From the perspective of indicators, our proposed method has higher accuracy, weighted recall, and weighted F1 than the baseline method. Specifically, the accuracy of our method is 64.5\%, which is higher than the accuracy of DvsGesture that is 63.4\%. As the accuracy is computed by the overall samples without considering the imbalance distribution of SGED, more in-depth and detailed analysis is needed. For weighted precision metric, the DvsGesture is better than ours, this phenomenon may caused by the DvsGesture's tendency to predict positive categories with more samples, which can be verified again by the confusion matrix later. Our method takes into account the imbalance of samples, so the performance indicators of the method proposed in this paper are better on indicators weighted recall and weighted F1.

\begin{figure}
	\centering
	\subfigure[]{\label{fig:subfig:a}
		\includegraphics[width=0.45\linewidth]{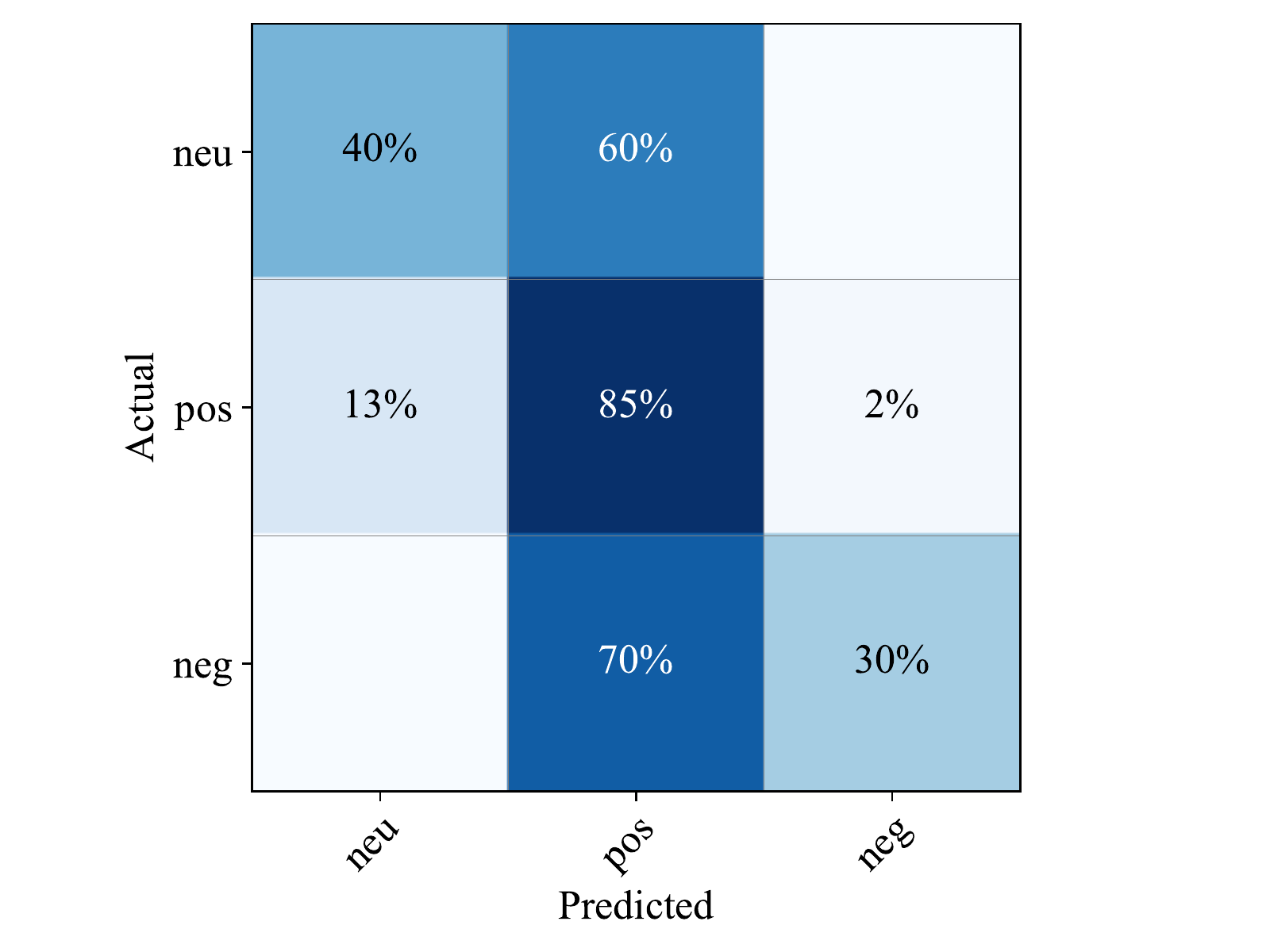}}
	\hspace{0.01\linewidth}
	\subfigure[]{\label{fig:subfig:b}
		\includegraphics[width=0.45\linewidth]{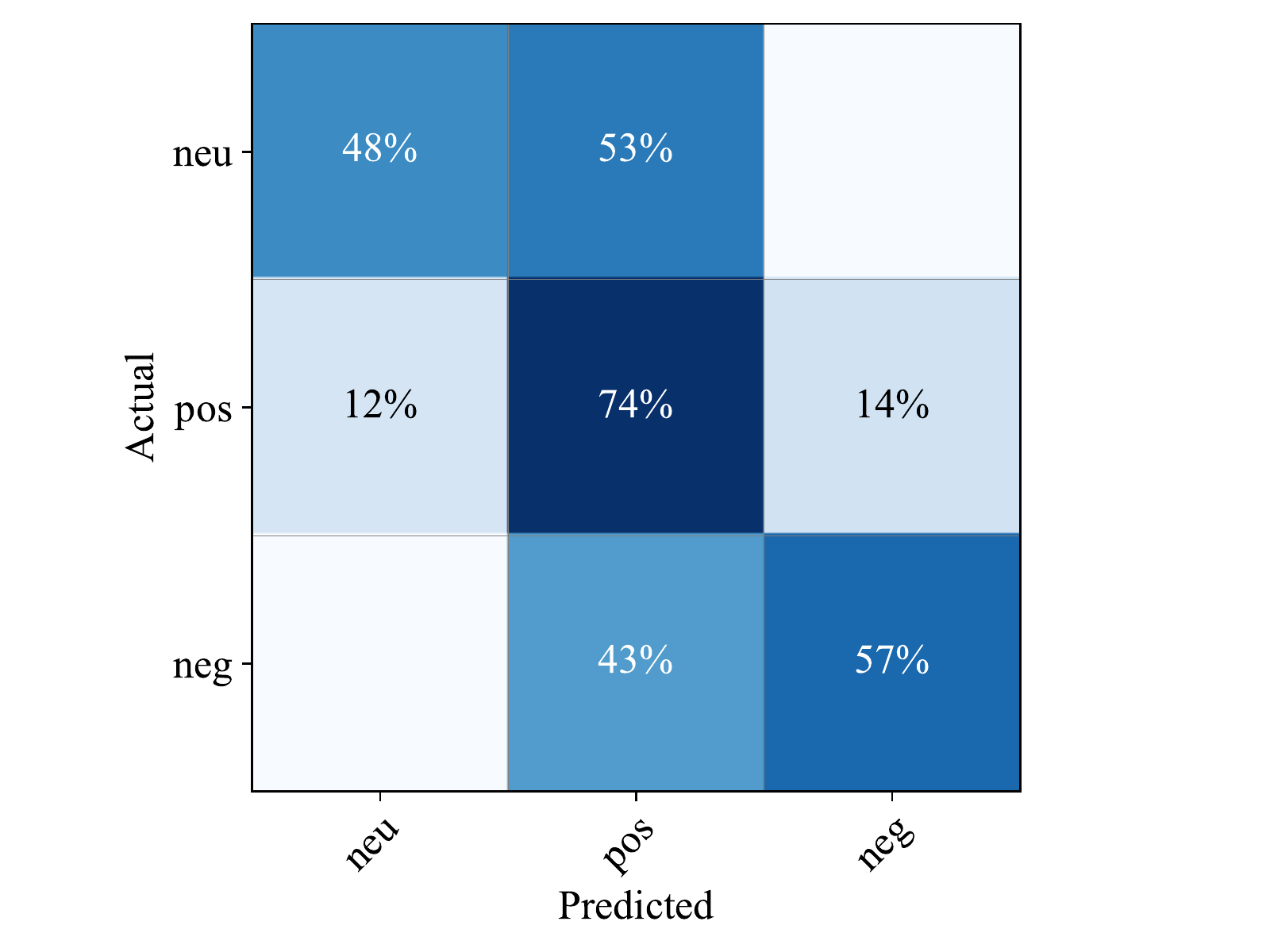}}
	\caption{(a) The confusion matrix of the results of the baseline method. (b) The confusion matrix of the method proposed in this paper.}
	\label{subfigs}
\end{figure}

To more intuitively see the distribution of different categories of predictions, Figure \ref{subfigs} shows the confusion matrix for the results of the two methods. From the confusion matrix, it can be seen that the advantage of the method proposed in this paper comes from the correct prediction of neutral and negative gesture emotions. For positive gesture emotions, our performance is not comparable to the baseline method. This reason is attributed to the uneven distribution of gesture emotion categories. Although we considered the factors of category imbalance in the design process of the loss function, there are still models that tend to output positive gesture emotion results. We believe that this is also one of the challenging features of the dataset released in this paper.

\section{Conclusion and Future Work}
\label{sec:5}
This paper proposes a homogeneous multimodal gesture emotion recognition dataset that includes event stream data and video frame data. Furthermore, a pseudo dual stream structured network is proposed to provide a benchmark solution. This dataset is a combination of the field of affective computing and the research field of gesture recognition. For affective computing, homogeneous multimodal data can be added to existing multimodal data, enriching data diversity, and leveraging the high dynamic range characteristics of event cameras. For spiking neural networks, the homogenous modal data we provide synchronously can promote the research of the spiking neurons' encoding ability for gesture emotion information. The future research goal can be to improve the fusion mechanism of two homogeneous multimodal data, thereby ultimately improving the accuracy of gesture emotion recognition. In addition, deploying on neuromorphic hardware to achieve low-power solutions is also one of the future research directions \cite{davies2018loihi}.

\section{Acknowledgments}
This work was supported by the Natural Science Foundation of Shandong Province (No. ZR2021QF145). We would like to thank all those who have contributed to the process of dataset construction. As the first author, I would like to thank my precious sun, Wenshi Wang, for coming to my life.


\bibliography{emotion}

\end{document}